%% file: main.tex
\documentclass[final]{cvpr}
\usepackage[dvipsnames,table]{xcolor}
\usepackage{times}
\usepackage{epsfig}
\usepackage{graphicx}        
\usepackage{amsmath}
\usepackage{amssymb}
\usepackage{color}
\usepackage{tabularx}
\usepackage{booktabs}
\usepackage{multirow}
\usepackage[linesnumbered,ruled]{algorithm2e}
\usepackage{array}
\usepackage{xspace}
\usepackage{balance}
\usepackage{enumitem}
\usepackage{mathtools}
\usepackage{rotating}
\usepackage{url}
\usepackage{tabulary}
\usepackage{subcaption}
\usepackage[export]{adjustbox}
\usepackage{ctable}
\usepackage{diagbox}
\usepackage{subcaption}
\usepackage{bbm}
\usepackage{makecell}
\usepackage{amssymb}
\usepackage{url}
\usepackage[title]{appendix}

\usepackage{dblfloatfix}

\graphicspath{{./figs/}}

\usepackage[pagebackref=true,breaklinks=true,colorlinks,bookmarks=false]{hyperref}

\begin{document}

	\title{CoMatch: Semi-supervised Learning with Contrastive Graph Regularization}
	
	\author{Junnan Li~~~Caiming Xiong~~~Steven C.H. Hoi\\
		Salesforce Research\\
	{\tt\small \{junnan.li,cxiong,shoi\}@salesforce.com}		
	}

	\maketitle
	
	\input{sec_abstract}
	\input{sec_introduction}

	\input{sec_literature}

	\input{sec_method}
	\input{sec_experiment}
	\input{sec_conclusion}

	{\small
		\bibliographystyle{ieee_fullname}
		\bibliography{bib}
	}
\input{sec_appendix}

\end{document}

%% file: sec_abstract.tex
\begin{abstract}
Semi-supervised learning has been an effective paradigm for leveraging unlabeled data to reduce the reliance on labeled data.
We propose CoMatch,
a new semi-supervised learning method that unifies dominant approaches and addresses their limitations.
CoMatch jointly learns two representations of the training data, their class probabilities and low-dimensional embeddings.
The two representations interact with each other to jointly evolve.
The embeddings impose a smoothness constraint on the class probabilities to improve the pseudo-labels,
whereas the pseudo-labels regularize the structure of the embeddings through graph-based contrastive learning.
CoMatch achieves state-of-the-art performance on multiple datasets.
It achieves substantial accuracy improvements on the label-scarce CIFAR-10 and STL-10.
On ImageNet with 1\% labels,
CoMatch achieves a top-1 accuracy of 66.0\%,
outperforming FixMatch~\cite{fixmatch} by 12.6\%.
Furthermore,
CoMatch achieves better representation learning performance on downstream tasks,
outperforming both supervised learning and self-supervised learning.
Code and pre-trained models are available at 
\textcolor{magenta}{\url{https://github.com/salesforce/CoMatch/}}.
\vspace{-2ex}
\end{abstract}

%% file: sec_introduction.tex
\vspace{-2ex}
\section{Introduction}
\label{sec:introduction}
\vspace{-0.5ex}

Semi-supervised learning (SSL)~\textendash~learning from few labeled data and a large amount of unlabeled data~\textendash~has been a long-standing problem in computer vision and machine learning.
Recent state-of-the-art methods mostly follow two trends:
(1) using the model's class prediction to produce a pseudo-label for each unlabeled sample as the label to train against~\cite{pseudo,mixmatch, remixmatch, fixmatch};
(2) unsupervised or self-supervised pre-training, followed by supervised fine-tuning~\cite{simclr,moco,byol,swav} and pseudo-labeling~\cite{bigsemi}.

However,
existing methods have several limitations.
Pseudo-labeling (also called self-training) methods heavily rely on the quality of the model's class prediction,
thus suffering from confirmation bias where the prediction mistakes would accumulate.
Self-supervised learning methods are task-agnostic,
and the widely adopted contrastive learning~\cite{simclr,moco} may learn representations that are suboptimal for the specific classification task.
Another branch of methods explore graph-based semi-supervised learning~\cite{teacher_graph,deep_label_prop},
but have yet shown competitive performance especially on larger datasets such as ImageNet~\cite{imagenet}.

We propose CoMatch, a new semi-supervised learning method that addresses the existing limitations.
A conceptual illustration is shown in Figure~\ref{fig:overview}.
In CoMatch,
each image has two compact representations:
a class probability produced by the classification head and a low-dimensional embedding produced by the projection head.
The two representations interact with each other and jointly evolve in a co-training framework.
Specifically,
the classification head is trained using memory-smoothed pseudo-labels,
where pseudo-labels are refined by aggregating information from nearby samples in the embedding space. 
The projection head is trained using contrastive learning on a pseudo-label graph,
where samples with similar pseudo-labels are trained to have similar embeddings. 
CoMatch unifies dominant ideas including consistency regularization, 
entropy minimization,
contrastive learning, and graph-based SSL.

\input{table/fig_overview.tex}

We perform experiments on multiple datasets and compare with state-of-the-art semi-supervised and self-supervised methods.
CoMatch substantially outperforms all baselines across all benchmarks,
especially in label-scarce scenarios.
On CIFAR-10 with 4 labeled samples per class, CoMatch outperforms FixMatch~\cite{fixmatch} by 6.11\% in accuracy.
On STL-10, CoMatch outperforms FixMatch by 13.27\%.
On ImageNet with only 1\% of labels, CoMatch achieves a top-1 accuracy of 66.0\% (67.1\% with self-supervised pre-training),
whereas the best baseline (MoCov2~\cite{mocov2} followed by FixMatch~\cite{fixmatch}) has an accuracy of 59.9\%.
Furthermore,
we demonstrate that CoMatch achieves better representation learning performance on down-stream image classification and object detection tasks,
outperforming both supervised learning and self-supervised learning.


%% file: table/fig_overview.tex
\begin{figure*}[!t]

	\centering
	\includegraphics[width=\textwidth]{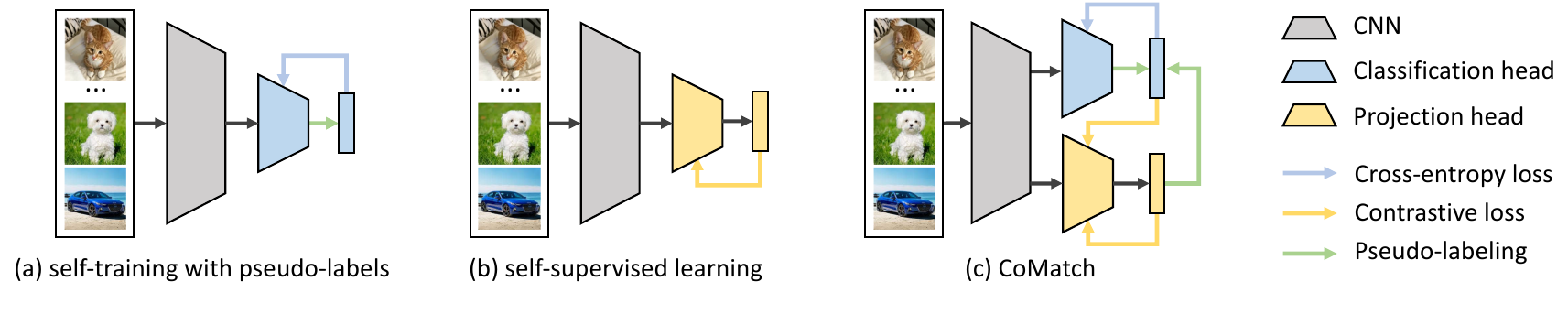}
  \vspace{-6ex}
  \caption
  	{ \small
	Conceptual illustration of different methods that leverage unlabeled data.
	(a) Task-specific self-training: the model predicts class probabilities for the unlabeled samples as the pseudo-label to train against~\cite{pseudo, mixmatch,remixmatch, fixmatch}. 
	(b) Task-agnostic self-supervised learning: the model projects samples into low-dimensional embeddings, and performs contrastive learning to discriminate embeddings of different images~\cite{instance,simclr,moco}.
	(c) CoMatch: class probabilities and embeddings interact with each other and jointly evolve in a co-training framework. 
	The embeddings impose a smoothness constraint on the class probabilities to improve the pseudo-labels.
	The pseudo-labels are used as the target to train both the classification head with a cross-entropy loss, and the projection head with a graph-based contrastive loss.
	  } 
    \vspace{-1ex}
  \label{fig:overview}

 \end{figure*} 

%% file: sec_literature.tex
\vspace{-1ex}
\section{Background}
\label{sec:literature}
\vspace{-0.5ex}
To set the stage for CoMatch,
we first introduce existing SSL methods,
mainly focusing on current state-of-the-art methods that are relevant.
More comprehensive reviews can be found in~\cite{survey_2005,survey_2020}. 
In the following,
we refer to a deep encoder network (a convolutional neural network) as $f(\cdot)$,
which produces a high-dimensional feature $f(x)$ given an input image $x$.
A classification head (a fully-connected layer followed by softmax) is defined as $h(\cdot)$,
which outputs a distribution over classes $p(y|x)=h(f(x))$.
We also define a non-linear projection head (a MLP) $g(\cdot)$,
which transforms a feature $f(x)$ into a normalized low-dimensional embedding $z(x)=g(f(x))$.

\vspace{0.5ex}
\noindent\textbf{Consistency regularization} is a crucial piece for many state-of-the-art SSL methods.
It utilizes the assumption that a classifier should output the same class probability for an unlabeled sample even after it is augmented.
In the simplest form,
prior works~\cite{consistency,temporal} add the following consistency regularization loss on unlabeled samples:
\begin{equation}
\label{eqn:consistency}
\left\lVert {p(y|\mathrm{Aug}(x))-p(y|\mathrm{Aug}(x))} \right\rVert^2_2,
\end{equation}
where $\mathrm{Aug}(\cdot)$ is a stochastic transformation that does not alter the label of the image.
Mean Teacher~\cite{mean_teacher} replaces one of the terms in eq.(\ref{eqn:consistency}) with the output of an EMA model.
VAT~\cite{VAT} uses an adversarial transformation in place of $\mathrm{Aug}$.
MixMatch~\cite{mixmatch} averages predictions across multiple augmentations to produce $p(y)$.
UDA~\cite{UDA}, ReMixMatch~\cite{remixmatch}, and FixMatch~\cite{fixmatch} use a cross-entropy loss in place of the squared error, and apply stronger augmentation. 

\vspace{0.5ex}
\noindent\textbf{Entropy minimization} is a common method in many SSL algorithms,
which encourages the classifier's decision boundary to pass through low-density regions of the data distribution.
It is either achieved explicitly by minimizing the entropy of $p(y|x)$ on unlabeled samples~\cite{EntMin},
or implicitly by constructing low-entropy \textbf{pseudo-labels} on unlabeled samples and using them as training targets in a cross-entropy loss~\cite{pseudo,mixmatch,remixmatch,fixmatch}.
Some methods~\cite{UDA,mixmatch,remixmatch} post-process the ``soft'' pseudo-labels with a sharpening function to reduce entropy,
whereas FixMatch~\cite{fixmatch} produces ``hard'' pseudo-labels for samples whose largest class probability fall above a predefined threshold.   
Most methods~\cite{fixmatch,remixmatch,UDA} use weakly-augmented samples to produce pseudo-labels and train the model on strongly-augmented samples.
However, 
since the pseudo-labels purely rely on the classifier,
such self-training strategy suffers from the confirmation bias problem,
where the error in the pseudo-labels would accumulate and harms learning.


\vspace{0.5ex}
\noindent\textbf{Self-supervised contrastive learning}
has attracted much attention,
due to its ability to leverage unlabeled data for model pre-training.
The widely adopted contrastive learning~\cite{instance,CPC,simclr,bigsemi,moco} optimizes for the task of instance discrimination,
and formulates the loss using the normalized low-dimensional embeddings $z$:
 \begin{equation}
 \label{eqn:contrastive}
 -\log\frac{\exp(z(\mathrm{Aug}(x_i))\cdot z(\mathrm{Aug}(x_i))/t)}{\sum_{j=1}^N\exp(z(\mathrm{Aug}(x_i))\cdot z(\mathrm{Aug}(x_j))/t)}
 \end{equation}
where $\mathrm{Aug}(\cdot)$ is a stochastic transformation similar as in eq.(\ref{eqn:consistency}),
and $x_j$ include $x_i$ and $N-1$ other images (\ie~negative samples).
Self-supervised contrastive learning can be interpreted as a form of class-agnostic consistency regularization,
which enforces the same image with different augmentations to have similar embeddings,
while different images have different embeddings.
Among recent methods,
SimCLR~\cite{simclr} uses images from the same batch to calculate pairwise similarity,
whereas MoCo~\cite{moco} maintains a queue of embeddings from an EMA model.

Self-supervised pre-training followed by supervised fine-tuning has shown strong performance on semi-supervised learning tasks~\cite{simclr,moco,byol,pcl,swav}.
SimCLR v2~\cite{bigsemi} further utilizes larger models for distillation.
However,
since self-supervised learning is a task-agnostic process,
the contrastive loss in eq.(\ref{eqn:contrastive}) optimizes for an objective that partially contradicts with task-specific learning.
It enforces images from the same class to have different representations,
which is undesirable for classification tasks.

\vspace{0.5ex}
\noindent\textbf{Graph-based semi-supervised learning} defines the similarity of data samples with a graph and encourages smooth predictions with respect to the graph structure~\cite{label_prop,Zhu_ICML_2013}.
Recent works use deep networks to generate graph representations.
\cite{deep_label_prop,dense_label_prop} perform iterative label propagation and network training.
\cite{teacher_graph,rem} connect data samples that have the same pseudo-labels,
and perform metric learning to enforce connected samples to have similar representations.
However,
these methods define representations as the high-dimensional feature $f(x)$,
which leads to several limitations: (1) since the features are highly-correlated with the class predictions, the same types of errors are likely to exist in both the feature space and the label space;
(2) due to the curse of dimensionality, Euclidean distance becomes less meaningful;
(3) computation cost is high which harms the scalability of the methods.
Furthermore, the loss functions in~\cite{teacher_graph,rem} consider the absolute distance between pairs,
whereas CoMatch optimizes for relative distance.

%% file: sec_method.tex
\section{Method}
\label{sec:method}
\subsection{Overview}

In this section, we introduce our proposed semi-supervised learning method.
Different from most existing semi-supervised and self-supervised learning methods,
CoMatch jointly learns the encoder $f(\cdot)$,
the classification head $h(\cdot)$,
and the projection head $g(\cdot)$.
Given a batch of $B$ labeled samples $\mathcal{X}=\{(x_b,y_b)\}_{b=1}^B$ where $y_b$ are one-hot labels,
and a batch of unlabeled samples $\mathcal{U}=\{u_b\}_{b=1}^{\mu B}$ where $\mu$ determines the relative size of $\mathcal{X}$ and $\mathcal{U}$,
CoMatch jointly optimizes three losses:
(1) a supervised classification loss on labeled data $\mathcal{L}_x$,
(2) an unsupervised classification loss on unlabeled data $\mathcal{L}_u^{cls}$,
and (3) a graph-based contrastive loss on unlabeled data $\mathcal{L}_u^{ctr}$.
Specifically, $\mathcal{L}_x$ is defined as the cross-entropy between the ground-truth labels and the model's predictions:
\begin{equation}
	\mathcal{L}_x = \frac{1}{B} \sum_{b=1}^{B} \mathrm{H}(y_b,p(y|\mathrm{Aug_w}(x_b))),
\end{equation}
where $\mathrm{H}(y,p)$ denotes the cross-entropy between two distributions $y$ and $p$, and $\mathrm{Aug_w}$ refers to weak augmentations.

The unsupervised classification loss $\mathcal{L}_u^{cls}$ is defined as the cross-entropy between the pseudo-labels $q_b$ and the model's predictions: 
\begin{equation}
\label{eqn:u_cls}
\mathcal{L}_u^{cls} = \frac{1}{\mu B} \sum_{b=1}^{\mu B} \mathbbm{1} (\max q_b\geq \tau) \mathrm{H}(q_b,p(y|\mathrm{Aug_s}(u_b))),
\end{equation}
where $\mathrm{Aug_s}$ refers to strong augmentations.
Following FixMatch~\cite{fixmatch},
we retain pseudo-labels whose largest class probability are above a threshold $\tau$.
Different from FixMatch, our soft pseudo-labels $q_b$ are not converted to hard labels for entropy minimization.
Instead, we achieve entropy minimization by optimizing the contrastive loss $\mathcal{L}_u^{ctr}$.
Section~\ref{sec:comatch} explains the details of pseudo-labelling and contrastive learning.

Our overall training objective is:
\begin{equation}
\mathcal{L}=\mathcal{L}_x + \lambda_{cls} \mathcal{L}_u^{cls} + \lambda_{ctr} \mathcal{L}_u^{ctr},
\end{equation}
where $\lambda_{cls}$ and $\lambda_{ctr}$ are scalar hyperparameters to control the weight of the unsupervised losses.

\input{table/fig_framework}

\subsection{CoMatch}
\label{sec:comatch}
In CoMatch,
the high-dimensional feature of each sample is transformed to two compact representations:
its class probability $p$
and its normalized low-dimensional embedding $z$,
which reside in the label space and the embedding space, respectively.
Given a batch of unlabeled samples $\mathcal{U}$,
we first perform memory-smoothed pseudo-labeling on weak augmentations $\mathrm{Aug_w}(\mathcal{U)}$ to produce pseudo-labels.
Then, we construct a pseudo-label graph $W^q$ which defines the similarity of samples in the label space.
We use $W^q$ as the target to train an embedding graph $W^z$,
which measures the similarity of strongly-augmented samples $\mathrm{Aug_s}(\mathcal{U)}$ in the embedding space.
An illustration of CoMatch is shown in Fig~\ref{fig:framework},
and a pseudo-code is given in the appendix.
Next, we first introduce the pseudo-labeling process,
then we describe the graph-based contrastive learning algorithm.

\vspace{0.5ex}
\noindent\textbf{Memory-smoothed pseudo-labeling} aims to mitigate confirmation bias by leveraging the structure of the embeddings to refine pseudo-labels.
Given each sample in $\mathcal{X}$ and $\mathcal{U}$,
we first obtain its class probability.
For a labeled sample, it is defined as the ground-truth label:
$p^w=y$.
For an unlabeled sample, it is defined as the model's prediction on its weak-augmentation:
$p^w=h\circ f(\mathrm{Aug_w}(u))$.
Following~\cite{remixmatch},
we perform distribution alignment (DA) on unlabeled samples:
$p^w=\mathrm{DA}(p^w)$.
DA prevents the model's prediction from collapsing to certain classes. 
Specifically,
we maintain a moving-average $\tilde{p}^w$ of $p^w$ during training,
and adjust the current $p^w$ with $p^w = \mathrm{Normalize}(p^w/\tilde{p}^w)$,
where $\mathrm{Normalize}(p)_i=p_i / \sum_j p_j$ renormalizes the scaled result to a valid probability distribution.

For each sample in $\mathcal{X}$ and $\mathcal{U}$, we also obtain its embedding $z^w$ by forwarding the weakly-augmented sample through $f$ and $g$.
Then, we create a memory bank to store class probabilities and embeddings of the past $K$ weakly-augmented samples:
$\mathrm{MB}=\{(p_k^w,z_k^w)\}_{k=1}^K$.
The memory bank contains both labeled samples and unlabeled samples and is updated with first-in-first-out strategy.



For each unlabeled sample $u_b$ in the current batch with $p^w_b$ and $z^w_b$,
we generate a pseudo-label $q_b$ by aggregating class probabilities from neighboring samples in the memory bank. 
Specifically,
we optimize the following objective:
\begin{equation}
	J(q_b) = (1-\alpha) \sum_{k=1}^K a_k \left\lVert q_b-p^w_k \right\rVert_2^2 + \alpha \left\lVert q_b-p^w_b \right\rVert_2^2 \big.
\end{equation}
The first term is a smoothness constraint which encourages $q_b$ to take a similar value as its nearby samples' class probabilities,
whereas the second term attempts to maintain its original class prediction.
$a_k$ measures the affinity between the current sample and the $k$-th sample in the memory,
and is computed using similarity in the embedding space:
\vspace{-0.5ex}
\begin{equation}
a_k = \frac{\exp({z^w_b}\cdot{z^w_k}/t)}{\sum_{k=1}^{K}\exp({z^w_b}\cdot{z^w_k}/t)},
\end{equation}
\vspace{-0.5ex}
where $t$ is a scalar temperature parameter.

Since $a_k$ is normalized (\ie~$a_k$ sums to one), the minimizer for $J(q_b)$ can be derived as:
\vspace{-0.8ex}
\begin{equation}
\label{eqn:pseudo_label}
q_b  = \alpha  p^w_b + (1-\alpha) \sum_{k=1}^K a_k p^w_k.
\end{equation}
\vspace{-0.8ex}

\noindent\textbf{Graph-based contrastive learning} aims to learn representations guided by a pseudo-label graph. 
Given the pseudo-labels $\{q_b\}_{b=1}^{\mu B}$ for the batch of unlabeled samples,
we build the pseudo-label graph by constructing a similarity matrix $W^q$ of size $\mu B \times \mu B$:
\vspace{-0.5ex}
\begin{equation}
\label{eqn:graph_q}
W^q_{bj} = 
\begin{cases}
1 & \text{if}~b=j\\
q_b \cdot q_j & \text{if}~b\neq j \text{~and~} q_b \cdot q_j\geq T\\
0 & \text{otherwise}
\end{cases}
\end{equation}
Samples with similarity lower than a threshold $T$ are not connected,
and each sample is connected to itself with the strongest edge of value 1 (\ie self-loop).

The pseudo-label graph serves as the target to train an embedding graph.
To construct the embedding graph,
we first perform two strong augmentations on each unlabeled sample $u_b\in \mathcal{U}$,
and obtain their embeddings $z_b=g\circ f(\mathrm{Aug_s}(u_b))$, $z'_b=g\circ f(\mathrm{Aug'_s}(u_b))$.
Then we build the embedding graph $W^z$ as:
\vspace{-0.5ex}
\begin{equation}
W^z_{bj} = 
\begin{cases}
\exp(z_b \cdot z'_b/t)& \text{if}~b=j\\
\exp(z_b \cdot z_j/t)& \text{if}~b\neq j \\
\end{cases}
\end{equation}
\vspace{-0.5ex}

We aim to train the encoder $f$ and the projection head $g$ such that the embedding graph has the same structure as the pseudo-label graph.
To this end,
we first normalize $W^q$ and $W^z$ with $\hat{W}_{bj}=W_{bj} / \sum_j W_{bj}$, so that each row of the similarity matrix sums to 1.
Then we minimize the cross-entropy between the two normalized graphs.
The contrastive loss is defined as:

\vspace{-1ex}
\begin{equation}
	\mathcal{L}_u^{ctr} = \frac{1}{\mu B} \sum_{b=1}^{\mu B} \mathrm{H}(\hat{W^q_b},\hat{W^z_b})
\end{equation}
\vspace{-0.5ex}
$\mathrm{H}(\hat{W^q_b},\hat{W^z_b})$ can be decomposed into two terms:
\vspace{-0.5ex}
\begin{equation}
  \resizebox{1\hsize}{!}{$
- \hat{W}^q_{bb}  \log (\frac{\exp(z_b \cdot z'_b/t)}{\sum_{j=1}^{\mu B}\hat{W}^z_{bj}} ) - \sum\limits_{j=1,j\neq b}^{\mu B}\hat{W}^q_{bj}  \log (\frac{\exp(z_b \cdot z_j/t)}{\sum_{j=1}^{\mu B}\hat{W}^z_{bj}} )	
	$}
\end{equation}
The first term is a self-supervised contrastive loss that comes from the self-loops in the pseudo-label graph.
It encourages the model to produce similar embeddings for different augmentations of the same image,
which is a form of consistency regularization.
The second term encourages samples with similar pseudo-labels to have similar embeddings.
It gathers samples from the same class into clusters, 
which achieves entropy minimization. 

During training,
a natural curriculum would occur from CoMatch.
The model would start with producing low-confidence pseudo-labels,
which leads to a sparse pseudo-label graph.
As training progresses,
samples are gradually clustered, 
which in turns leads to more confident pseudo-labels and more connections in the pseudo-label graph.

Another advantage of CoMatch appears in open-set semi-supervised learning,
where the unlabeled data contains out-of-distribution (ood) samples.
Due to the smoothness constraint,
ood samples would have low-confidence pseudo-labels.
Therefore, they are less connected to in-distribution samples,
and will be pushed further away from in-distribution samples by the proposed contrastive loss.


\vspace{-0.5ex}
\subsection{Scalable learning with an EMA model}
\vspace{-0.5ex}
In order to build a meaningful pseudo-label graph,
the unlabeled batch of data should contain a sufficient number of samples from each class.
While this requirement can be easily satisfied for datasets with a small number of classes (\eg~CIFAR-10),
it becomes difficult for large datasets with more classes (\eg~ImageNet) because a large unlabeled batch would exceed the memory capacity of 8 commodity GPUs (\eg NVIDIA~V100).
Therefore,
we improve CoMatch for SSL on large-scale datasets. 

Inspired by MoCo~\cite{moco} and Mean Teacher~\cite{mean_teacher},
we introduce an EMA model $\{\bar{f},\bar{g},\bar{h}\}$ whose parameters $\bar{\theta}$ are the moving-average of the original model's parameters $\theta$:
\vspace{-0.5ex}
\begin{equation}
\bar{\theta} \leftarrow m \bar{\theta} + (1-m) \theta.
\end{equation}
The advantage of the EMA model is that it can evolve smoothly as controlled by the momentum parameter $m$.

We also introduce a momentum queue
which stores the pseudo-labels and the strongly-augmented embeddings for the past $K$ unlabeled samples:
$\mathrm{MQ}=\{(\bar{q}_k,\bar{z}_k=\bar{g}\circ \bar{f}(\mathrm{Aug_s}'(u_k)))\}_{k=1}^K$,
where $\bar{q}_k$ and $\bar{z}_k$ are produced using the EMA model.  
Different from the memory bank,
the momentum queue only contains unlabeled samples.

We modify the pseudo-label graph $W^q$ to have a size of $\mu B \times K$.
It defines the similarity between each sample in the current batch and each sample in the momentum queue (which also contains the current batch).
Different from eqn.(\ref{eqn:graph_q}), the similarity is now calculated as $\bar{q}_b \cdot \bar{q}_j$,
where $b=\{1,...,\mu B\}$ and $j=\{1,...,K\}$.

The embedding graph $W^z$ is also modified to have a size of $\mu B \times K$,
where the similarity is calculated using the model's output embedding $z_b$ and the momentum embedding $\bar{z}_j$:
$W_{bj}^z=\exp(z_b \cdot \bar{z}_j / t)$.
Since gradient only flows back through $z_b$,
we can use a large $K$ with only a small increase in GPU memory usage and computation time.

Besides the contrastive loss,
we also leverage the EMA model for memory-smoothed pseudo-labeling,
by forwarding the weakly-augmented samples through the EMA model instead of the original model.
A graphical illustration of the memory bank and the momentum queue is given in the appendix.


%
%
%


%% file: table/fig_framework.tex
\begin{figure*}[!t]

	\centering
	\includegraphics[width=\textwidth]{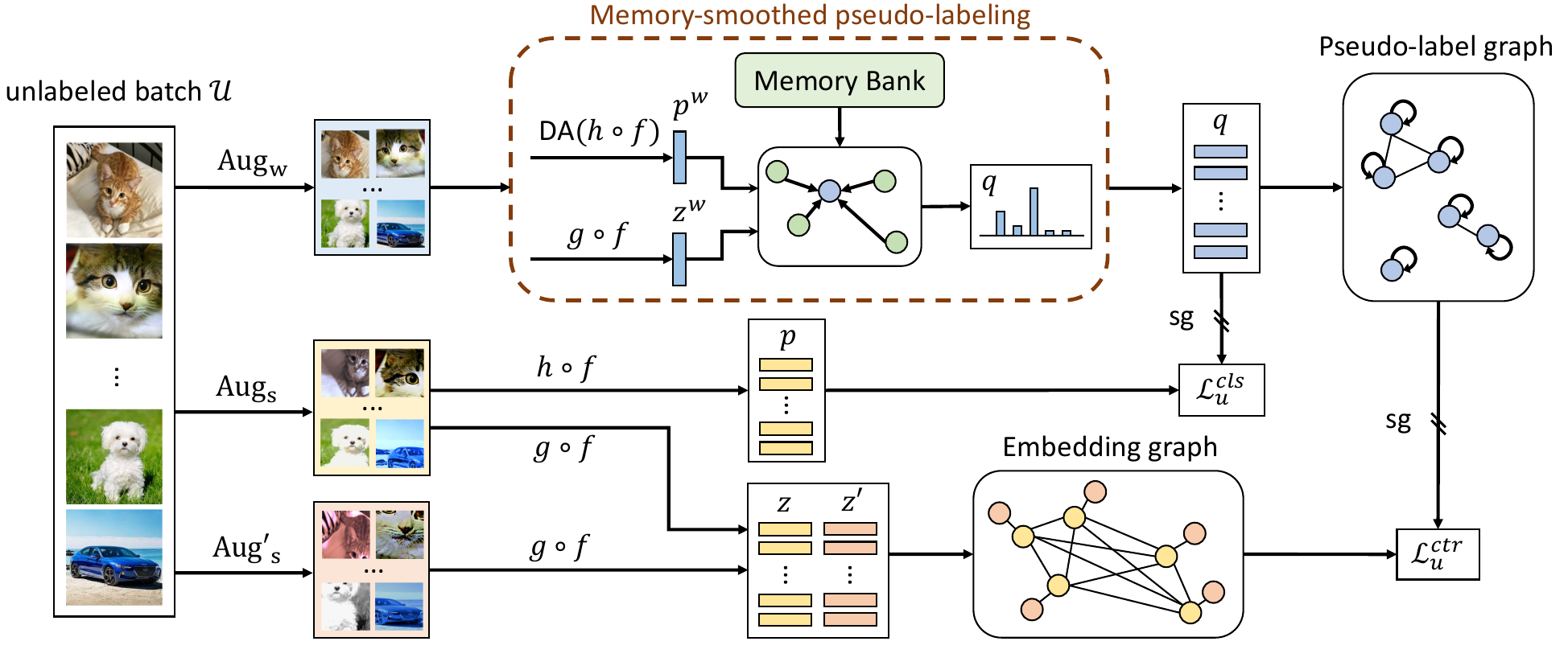}
  \vspace{-5ex}
  \caption
  	{ \small
		Framework of the proposed CoMatch. Given a batch of unlabeled images, their weakly-augmented images are used to produce memory-smoothed pseudo-labels, which are used as targets to train the class prediction on strongly-augmented images.
		A pseudo-label graph with self-loop is constructed to measure the similarity between samples, which is used to train an embedding graph such that images with similar pseudo-labels have similar embeddings. sg means stop-gradient.
	  } 
  \label{fig:framework}
  \vspace{-1ex}
 \end{figure*} 

%% file: sec_experiment.tex
\vspace{-0.5ex}
\section{Experiment}
\label{sec:experiment}

\vspace{-0.5ex}
\subsection{CIFAR-10 and STL-10}
\vspace{-0.5ex}
First, 
we conduct experiments on CIFAR-10 and STL-10 datasets.
CIFAR-10 contains 50,000 images of size $32\times32$ from 10 classes. 
We vary the amount of labeled data and focus on the label-scarce scenario where few labels are available.
We evaluate on 5 runs with different random seeds.
STL-10 contains 5,000 labeled images of size $96\times96$ from 10 classes and 100,000 unlabeled images including ood samples.
We evaluate on the 5 pre-defined folds.
Following~\cite{mixmatch,fixmatch},
we report the performance of an EMA model.
\input{table/cifar_stl}

\input{table/imagenet}

\noindent\textbf{Baseline methods.}
For fair comparison, we improve the current state-of-the-art method FixMatch~\cite{fixmatch} with distribution alignment~\cite{remixmatch} to build a stronger baseline.
We also compare with the original FixMatch and MixMatch~\cite{mixmatch}.
We omit previous methods such as $\Pi$-model~\cite{pi_model}, Pseudo-Labeling~\cite{pseudo}, and Mean Teacher~\cite{mean_teacher} due to their poorer performance as reported in~\cite{fixmatch}.
Following~\cite{realistic}, we reimplemented the baselines and performed all experiments using the same model architecture,
the same codebase (PyTorch~\cite{pytorch}), and the same random seeds.


\noindent\textbf{Implementation details.}
For CIFAR-10, we use a Wide ResNet-28-2~\cite{wrn}.
For STL-10, we use a ResNet-18~\cite{resnet} due to its lower computation cost compared to the WRN-37-2 used in~\cite{fixmatch}\footnote{The forward-pass GFLOPs/image is 0.34 for ResNet-18 and 2.58 for WRN-37-2.
Compared to ResNet-18, WRN-37-2 takes $3\times$GPU memory and $7\times$training time per epoch.}.
The projection head is a 2-layer MLP which outputs 64-dimensional embeddings.
The models are trained using SGD with a momentum of 0.9 and a weight decay of 0.0005.
We follow the original papers~\cite{mixmatch,fixmatch} and train the baselines for 1024 epochs,
using an learning rate of 0.03 with a cosine decay schedule.
We train CoMatch for only 512 epochs to demonstrate its efficiency in learning.
For the hyperparameters in CoMatch that also exist in~\cite{fixmatch},
we follow~\cite{fixmatch} and set $\lambda_{cls}=1$, $\tau=0.95$, $\mu=7$, $B=64$.
For other hyperparameters,
we fix $\alpha=0.9$, $K=2560$, $t=0.2$, $T=0.8$, and $\lambda_{ctr}=1$ for all CIFAR-10 experiments, and only changes $\lambda_{ctr}$ to 5 for STL-10.

\noindent\textbf{Augmentations.}
CoMatch uses one ``weak'' augmentation $\mathrm{Aug_w}$,
and two ``strong'' augmentations $\mathrm{Aug_s}$ and $\mathrm{Aug'_s}$.
The weak augmentation for all experiments is the standard crop-and-flip.
For strong augmentations,
we follow~\cite{fixmatch} and uses RandAugment~\cite{randaug} as $\mathrm{Aug_s}$.
For $\mathrm{Aug'_s}$, we follow the augmentation strategy in SimCLR~\cite{simclr} which applies random color jittering and grayscale conversion.

\noindent\textbf{Results.}
Table~\ref{tbl:cifar_stl} shows the results. 
CoMatch outperforms the best baseline across all settings.
The improvement is more substantial when fewer labeled samples are available.
For example, CoMatch achieves an average accuracy of 93.09\% on CIFAR-10 with only 4 labels per class,
whereas FixMatch (w. DA) has a lower accuracy of 86.98\% and a larger variance. 
On STL-10,
CoMatch also improves FixMatch (w. DA) by 13.27\%.
\vspace{-0.5ex}
\subsection{ImageNet}

\input{table/fig_training}
\input{table/ablation}
\vspace{-0.5ex}
We evaluate CoMatch on ImageNet ILSVRC-2012 to verify its efficacy on large-scale datasets.
Following~\cite{S4L,simclr},
we randomly sample 1\% or 10\% of images with labels in a class-balanced way (13 or 128 samples per-class, respectively),
while the rest of images are unlabeled.
Our results are not sensitive to different random seeds hence we use a fixed random seed.

\noindent\textbf{Baseline methods.}
The baselines include (1) semi-supervised learning methods and (2) self-supervised pre-training followed by fine-tuning.
Furthermore, we construct a state-of-the-art baseline which combines FixMatch (w. DA) with self-supervised pre-training using MoCov2~\cite{mocov2} (pre-trained for 800 epochs).
Self-supervised methods require additional model parameters during training due to the projection network.
We count the number of training parameters as those that require gradient update.
We also report the performance of SimCLRv2~\cite{bigsemi}.
However, the best model from SimCLRv2 uses substantially (33$\times$) larger pre-trained teacher models to produce high-quality pseudo-labels for distillation.
Hence CoMatch should not be directly compared to SimCLRv2.

\noindent\textbf{Implementation details.}
We use a ResNet-50~\cite{resnet} model as the encoder.
Following~\cite{mocov2,simclr}, the projection head is a 2-layer MLP which outputs 128-dimensional embeddings.
We train the model using SGD with a momentum of 0.9 and a weight decay of 0.0001.
The learning rate is 0.1, which follows a cosine decay schedule for 400 epochs.
For models that are initialized with MoCov2,
we use a smaller learning rate of 0.03.
The momentum parameter is set as $m=0.996$.
Other hyperparameters are shown in appendix~\ref{sec:details}.
We use the same strong augmentation for $\mathrm{Aug_s}$ and $\mathrm{Aug'_s}$, which applies crop-and-flip followed by color distortion.
For fair comparison with baselines,
we report the original model's performance instead of the EMA model's.

\noindent\textbf{Results.}
Table~\ref{tbl:imagenet} shows the result, where CoMatch achieves state-of-the-art performance.
CoMatch obtains a top-1 accuracy of 66.0\% on 1\% of labels.
Compared to the the best baseline (MoCov2 followed by FixMatch w. DA),
CoMatch achieves 6.1\% improvement with $3\times$ less training time.
With the help of MoCov2 pre-training, 
the performance of CoMatch can further improve to 67.1\% on 1\% of labels,
and 73.7\% on 10\% of labels.
In Figure~\ref{fig:training},
we further show that CoMatch produces pseudo-labels that are more confident and accurate.
Pre-training with MoCov2 helps speed up the convergence rate.

\input{table/voc}
\input{table/coco}
\subsection{Ablation Study.}
We perform extensive ablation study to examine the effect of different components in CoMatch.
We use ImageNet with 1\% labels as the main experiment.
Due to the number of experiments in our ablation study,
we report the top-1 accuracy after training for 100 epochs,
where the default setting of CoMatch achieves 57.1\%.

\noindent\textbf{Graph connection threshold.}
The threshold $T$ in eqn.(\ref{eqn:graph_q}) controls the sparsity of edges in the pseudo-label graph.
Figure~\ref{fig:ablation}(a) presents the effect of $T$.
As $T$ increases, samples whose pseudo-labels have lower similarity are disconnected.
Hence their embeddings are pushed apart by our contrastive loss. 
When $T=1$, the proposed graph-based contrastive loss downgrades to the self-supervised loss in eqn.(\ref{eqn:contrastive}) where the only connections are the self-loops.
Using the self-supervised contrastive loss decreases the performance by 2.8\%.

\noindent\textbf{Contrastive loss weight.}
We vary the weight $\lambda_{ctr}$ for the contrastive loss $\mathcal{L}_u^{ctr}$ and report the result in Figure~\ref{fig:ablation}(b),
where $\lambda_{ctr}=10$ gives the best performance.
With 10\% of ImageNet labels, $\lambda_{ctr}=2$ yields better performance.
We find that in general,
fewer labeled samples require a larger $\lambda_{ctr}$ to strengthen the graph regularization.

\noindent\textbf{Prediction weight in pseudo-labels.} 
Our memory-smoothed pseudo-labeling uses $\alpha$ to control the balance between the EMA model's prediction and smoothness constraint.
Figure~\ref{fig:ablation}(c) shows its effect,
where $\alpha=0.9$ results in the best performance.
When $\alpha=1$, the pseudo-labels are purely generated by the EMA model,
which reduces to the Mean-Teacher~\cite{mean_teacher} method.
The accuracy decreases by 2.1\% due to confirmation bias.
When $\alpha<0.9$, the pseudo-labels are over-smoothed.
A potential improvement is to apply sharpening~\cite{mixmatch} to pseudo-labels with smaller $\alpha$, but is not studied here due to the need of an extra hyperparameter.

\noindent\textbf{Size of memory bank and momentum queue.}
$K$ controls both the size of the memory bank for pseudo-labeling
and the size of the momentum queue for contrastive learning.
A larger $K$ considers more samples to enforce a structural constraint on the label space and the embedding space.
As shown in Figure~\ref{fig:ablation}(d),
the performance increases as $K$ increases from 10k to 30k,
but plateaus afterwards.
We would also like to highlight that the memory bank and the momentum queue only introduce a small computation overhead because (1) low-dimensional embeddings are stored, (2) gradients are not computed \textit{w.r.t} to the embeddings.

\vspace{-0.5ex}
\subsection{Transfer of Learned Representations}
\vspace{-0.5ex}

We further evaluate the quality of the representations learned by CoMatch by transferring it to other tasks.
Following~\cite{benchmark, pcl},
We first perform linear classification on two datasets:  PASCAL VOC2007~\cite{voc} for object classification and Places205~\cite{places} for scene recognition. 
We train linear SVMs using fixed representations from ImageNet pre-trained models. 
We preprocess all images by resizing them to 256 pixels along the shorter side and taking a 224$\times$224 center crop. 
The SVMs are trained on the global average pooling features of ResNet-50.
To study the transferability of the representations in few-shot scenarios, 
we vary the number of samples per-class ($k$) in the downstream datasets.

Table~\ref{tbl:voc} shows the results.
We compare CoMatch with standard supervised learning on labeled ImageNet and self-supervised learning (MoCov2~\cite{mocov2} and SwAV~\cite{swav}) on unlabeled ImageNet.
CoMatch with 10\% labels achieves higher performance on both datasets.
It is interesting to observe that self-supervised learning methods do not perform well in few-shot transfer,
and only catch up with supervised learning when $k$ increases.

In Table~\ref{tbl:coco},
we also show that compared to supervised and self-supervised learning,
CoMatch learns a better CNN backbone for object detection and instance segmentation on COCO~\cite{coco}.
We follow the exact same setting as~\cite{moco} to fine-tune a Mask-RCNN model~\cite{mask_rcnn} for $1\times$ or $2\times$ schedule.

%% file: table/cifar_stl.tex

\begin{table*}[!t]
\small
	\centering
	\begin{tabular}	{l | c  c  c  c| c }
	\toprule
    \multirow{2}{*}{Method} & \multicolumn{4}{c|}{CIFAR-10} & STL-10 \\
	& 20 labels& 40 labels & 80 labels & 250 labels & 1000 labels \\
	\midrule
	 MixMatch~\cite{mixmatch} &27.84$\pm$10.63 & 51.90$\pm$11.76 & 80.79$\pm$1.28 & 88.97$\pm$0.85 &38.02$\pm$8.29\\
	FixMatch~\cite{fixmatch} &82.32$\pm$9.77 & 86.12$\pm$3.53& 92.06$\pm$0.88 & 94.90$\pm$0.67& 65.38$\pm$0.42 \\	 
	FixMatch~\cite{fixmatch} w. DA~\cite{remixmatch} &83.81$\pm$9.35 & 86.98$\pm$3.40& 92.29$\pm$0.86 & 94.95$\pm$0.66& 66.53$\pm$0.39 \\
	CoMatch & \cellcolor{blue!20}\textbf{87.67}$\pm$8.47&  \cellcolor{blue!20}\textbf{93.09}$\pm$1.39&\cellcolor{blue!20}\textbf{93.97}$\pm$0.62&\cellcolor{blue!20}\textbf{95.09}$\pm$0.33 &\cellcolor{blue!20}\textbf{79.80}$\pm$0.38\\

	\bottomrule
	\end{tabular}	
\vspace{-1ex}
	\caption
		{\small	
		Accuracy for CIFAR-10 and STL-10 on 5 different folds. All methods are tested using the same data and codebase.
		}
	\vspace{-3ex}
	\label{tbl:cifar_stl}
\end{table*}		

%% file: table/imagenet.tex
\begin{table*}[!b]
	\vspace{-1ex}
\small
	\centering
	\begin{tabular}	{l | l | l | l | c  c |c  c }
	\toprule
	\multirow{3}{*}{\shortstack[l]{Self-supervised\\Pre-training}} & \multirow{3}{*}{Method} & \multirow{3}{*}{\#Epochs} & \multirow{3}{*}{\shortstack[l]{\#Paramters\\ (train/test)}} & \multicolumn{2}{c|}{Top-1} & \multicolumn{2}{c}{Top-5} \\
	& & &    & \multicolumn{2}{c|}{Label fraction} & \multicolumn{2}{c}{Label fraction} \\
	 & & &    &  1\% & 10\%& 1\% & 10\%\\
	\midrule
	\multirow{8}{*}{None}  & Supervised baseline~\cite{S4L} & $\sim$20 & 25.6M / 25.6M & 25.4& 56.4  & 48.4 &80.4\\
	&Pseudo-label~\cite{pseudo,S4L} & $\sim$100 & 25.6M / 25.6M & - & - & 51.6&82.4\\ 
	&VAT+EntMin.~\cite{VAT,EntMin,S4L}& - & 25.6M / 25.6M & - & 68.8 & -&88.5\\ 
		& S4L-Rotation~\cite{S4L} & $\sim$200 & 25.6M / 25.6M & - & 53.4 & - &83.8\\ 
		& UDA (RandAug)~\cite{UDA} & - &  25.6M / 25.6M & - & 68.8 & - &88.5\\
	 & FixMatch (RandAug)~\cite{fixmatch} & $\sim$300 &  25.6M / 25.6M & - & 71.5 & - & 89.1\\ 
	& FixMatch w. DA&  $\sim$400  &  25.6M / 25.6M & 53.4& 70.8& 74.4& 89.0 \\ 
	& CoMatch& $\sim$400 &  30.0M / 25.6M &\cellcolor{blue!20}\textbf{66.0} & \cellcolor{blue!20}\textbf{73.6} &\cellcolor{blue!20}\textbf{86.4} & \cellcolor{blue!20}\textbf{91.6}\\ 
	\midrule
	PIRL~\cite{PIRL} & \multirow{5}{*}{Fine-tune}&$\sim$800  &  26.1M / 25.6M  & 30.7&60.4 &  57.2 & 83.8 \\ 
	PCL~\cite{pcl} & &$\sim$200  &  25.8M / 25.6M  & - & - & 75.3&85.6 \\ 
	SimCLR~\cite{simclr} & &$\sim$1000  &  30.0M / 25.6M  & 48.3& 65.6  & 75.5&87.8 \\ 
	BYOL~\cite{byol} & &$\sim$1000 &  37.1M / 25.6M  & 53.2&  68.8& 78.4&   89.0\\ 
	SwAV~\cite{swav} & &$\sim$800 &  30.4M / 25.6M  & 53.9& 70.2 &  78.5&   89.9\\ 
	\cmidrule{2-8}
	\multirow{3}{*}{MoCov2~\cite{mocov2}}  & Fine-tune&$\sim$800  &  30.0M / 25.6M  & 49.8 & 66.1& 77.2 & 87.9\\ 
	&FixMatch w. DA& $\sim$1200 & 30.0M / 25.6M & 59.9 & 72.2 & 79.8 &89.5\\ 	
	& CoMatch& $\sim$1200  &  30.0M / 25.6M & \cellcolor{blue!20}\textbf{67.1}&\cellcolor{blue!20}\textbf{73.7} &\cellcolor{blue!20}\textbf{87.1} &\cellcolor{blue!20}\textbf{91.4} \\ 	
	\midrule
	\multirow{2}{*}{SimCLRv2*~\cite{bigsemi}} & Fine-tune&  $\sim$800 & 34.2M / 29.8M & 57.9& 68.4 & 82.5&89.2\\ 
	& Fine-tune+Distillation&  $>$1200 & 829.2M / 29.8M & {\color{gray}73.9} & {\color{gray}77.5}& {\color{gray}91.5}&{\color{gray}93.4}\\
	\bottomrule
	\end{tabular}
\vspace{-0.5ex}
	\caption
		{\small	
		Accuracy for ImageNet with 1\% and 10\% of labeled examples.
		SimCLRv2*~\cite{bigsemi} uses larger models for training and test.
		}
	\vspace{-0.5ex}
	\label{tbl:imagenet}
\end{table*}

%% file: table/fig_training.tex
\begin{figure*}[!t]
 \centering
 \small
 \begin{minipage}{0.32\textwidth}
	\centering
	\includegraphics[trim=0 10 0 0,clip,width=\textwidth]{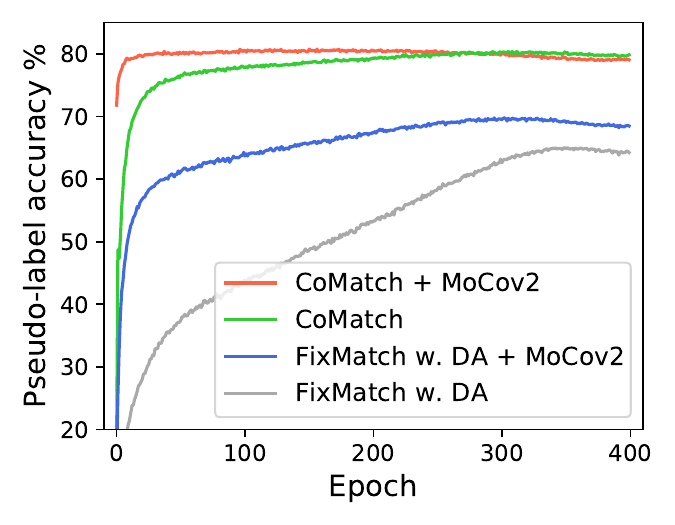}
	(a)
\end{minipage}
\hspace{0.2ex}
 \begin{minipage}{0.32\textwidth}
	\centering
	\includegraphics[trim=0 10 0 0,clip,width=\textwidth]{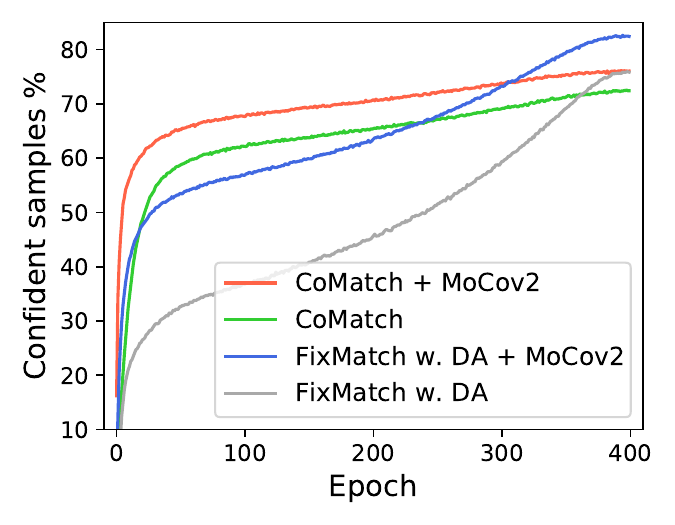}	
	(b)
\end{minipage}
\hspace{0.2ex}
 \begin{minipage}{0.32\textwidth}
	\centering
	\includegraphics[trim=0 10 0 0,clip,width=\textwidth]{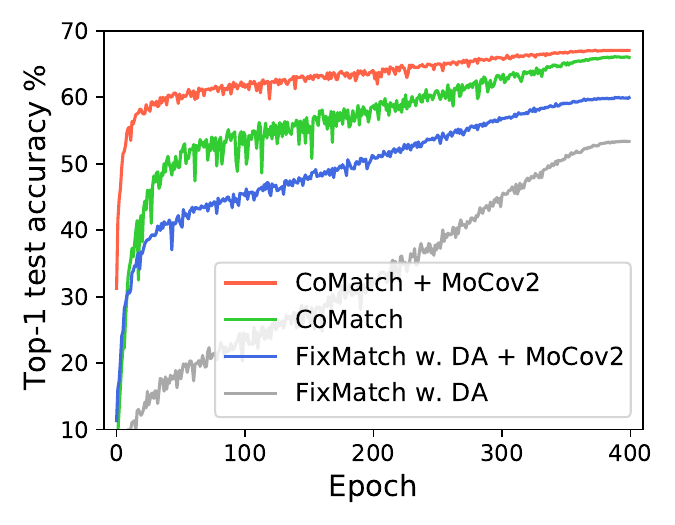}	
	(c)
\end{minipage}
\vspace{-2ex}
  \caption
  	{  	\small
  		Plots of different methods as training progresses on ImageNet with 1\% labels.
  		(a) Accuracy of the confident pseudo-labels \textit{w.r.t} to the ground-truth labels of the unlabeled samples.
  		(b) Ratio of the unlabeled samples with confident pseudo-labels that are included in the unsupervised classification loss. 
  		(3) Top-1 accuracy on the test data.
	  } 
  \vspace{-1ex}
  \label{fig:training}
 \end{figure*}

%% file: table/ablation.tex
\begin{figure*}[!t]
 \small
 \begin{minipage}{0.245\textwidth}
	\centering
	\includegraphics[trim=0 5 0 0,clip,width=\textwidth]{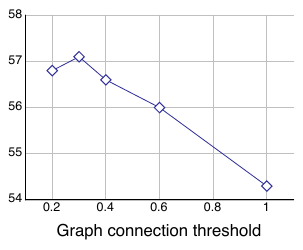}
	(a)
\end{minipage}
 \begin{minipage}{0.245\textwidth}
	\centering
	\includegraphics[trim=0 5 0 0,clip,width=\textwidth]{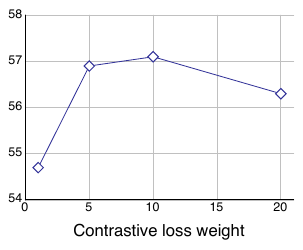}	
	(b)
\end{minipage}
 \begin{minipage}{0.245\textwidth}
	\centering
	\includegraphics[trim=0 5 0 0,clip,width=\textwidth]{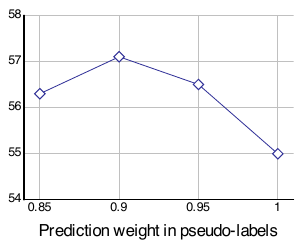}	
	(c)
\end{minipage}
 \begin{minipage}{0.245\textwidth}
	\centering
	\includegraphics[trim=0 5 0 0,clip,width=\textwidth]{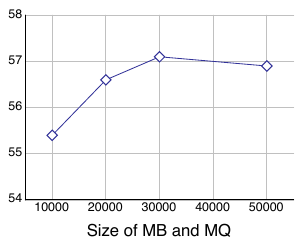}	
	(d)
\end{minipage}
\vspace{-2ex}
  \caption
  	{  	\small
  		Plots of various ablation studies on CoMatch. The default hyperparameter setting achieves 57.1\% (ImageNet with 1\% labels, trained for 100 epochs).
  		(a) Varying the threshold $T$ which controls the sparsity of edges in the pseudo-label graph. $T=1$ reduces to self-supervised contrastive learning.
  		(b) Varying the weight $\lambda_{ctr}$ for the contrastive loss.
  		(c) Varying $\alpha$, the weight of the EMA model's prediction in generating pseudo-labels.
  		$\alpha=1$ reduces to pseudo-labeling with mean teacher~\cite{mean_teacher}.
  		(d) Varying $K$, the number of samples in both the memory bank and the momentum queue. 
	  } 
  \label{fig:ablation}
  \vspace{-2ex}
 \end{figure*}

%% file: table/voc.tex
\begin{table*}[!t]
\small
	\centering
	\begin{tabular}	{l l l | c  c c c  c  }
	\toprule
    Method& \#ImageNet labels& \#Pre-train epochs & $k$=4 & $k$=8 &$k$=16 &$k$=64 & Full\\
	\midrule
	Supervised &  100\%  &90& 73.51$\pm$2.12& 79.60$\pm$0.61&82.75$\pm$0.34& 85.55$\pm$0.12& 87.12\\
	\midrule
	MoCov2~\cite{mocov2} &\multirow{3}{*}{0\%} &800 & 70.47$\pm$2.18& 76.74$\pm$0.87&80.61$\pm$0.53& 84.60$\pm$0.11& 86.83\\
	SwAV~\cite{swav} & &400 & 68.04$\pm$2.39& 75.06$\pm$0.73&79.46$\pm$0.55& 84.24$\pm$0.13& 86.86\\	
	\color{gray}{SwAV*~\cite{swav}} & &\color{gray}{800} & \color{gray}{64.27$\pm$2.13}& \color{gray}{73.19$\pm$0.68}&\color{gray}{78.87$\pm$0.46}& \color{gray}{85.07$\pm$0.20}& \color{gray}{88.10}\\
	\midrule
	CoMatch&1\%&400 & 72.81$\pm$1.50& 79.18$\pm$0.51&82.30$\pm$0.46& 85.65$\pm$0.17& 87.66\\
	CoMatch&10\%&400 & \textbf{74.56}$\pm$2.04& \textbf{80.60}$\pm$0.31&\textbf{83.24}$\pm$0.43& \textbf{86.07}$\pm$0.16&\textbf{87.91}\\	
	\bottomrule
	\end{tabular}	
\vspace{1ex}
\\(a) VOC07\\
\vspace{1.5ex}
	\begin{tabular}	{l l l | c  c c c  c  }
	\toprule
	Method & \#ImageNet labels& \#Pre-train epochs & $k$=4 & $k$=8 &$k$=16 &$k$=64 & $k$=256\\
	\midrule
	Supervised &  100\%  &90& 27.20$\pm$0.41& 32.08$\pm$0.45&35.95$\pm$0.21& 41.81$\pm$0.17& 45.74$\pm$0.14\\
	\midrule
	MoCov2~\cite{mocov2} &\multirow{3}{*}{0\%} &800 & 25.34$\pm$0.51& 30.64$\pm$0.39&35.08$\pm$0.34&42.18$\pm$0.10& 46.96$\pm$0.06\\
	SwAV~\cite{swav} & &400 & 25.32$\pm$0.46& 31.00$\pm$0.47&35.65$\pm$0.28& 42.60$\pm$0.11& \textbf{47.51}$\pm$0.20\\
	\color{gray}{SwAV*~\cite{swav}} & &\color{gray}{800} & \color{gray}{27.07$\pm$0.60}& \color{gray}{33.26$\pm$0.38}&\color{gray}{38.38$\pm$0.22}& \color{gray}{46.01$\pm$0.10}& \color{gray}{51.00$\pm$0.17}\\
	\midrule
	CoMatch&1\%&400 & 27.15$\pm$0.42& 32.36$\pm$0.37&36.56$\pm$0.33& 42.97$\pm$0.11& 47.32$\pm$0.18\\
	CoMatch&10\%&400 & \textbf{28.11}$\pm$0.33& \textbf{33.05}$\pm$0.46&\textbf{36.98}$\pm$0.28& \textbf{43.06}$\pm$0.22&47.10$\pm$0.11\\	
	\bottomrule
	
\end{tabular}	
\vspace{1ex}
\\(b) Places\\	
\vspace{-1ex}
	\caption
		{\small	
		Linear classification on VOC07 and Places using models pre-trained on ImageNet. We vary the number of examples per-class ($k$) on the down-stream datasets.
		We report the average result with std across 5 runs.
		SwAV* uses multi-crop augmentation.
		}
	\label{tbl:voc}
\end{table*}		

%% file: table/coco.tex
\begin{table*}[!t]
\small
	\centering	
	\begin{tabular}	{l c |  l l l  | l l l  |  l l l  | l l l }
	\toprule	 
	&\#ImageNet&\multicolumn{6}{c|}{$1\times$ schedule}&\multicolumn{6}{c}{$2\times$ schedule}	\\	
	Method & labels&AP$^\text{bb}$&   AP$^\text{bb}_{50}$ &  AP$^\text{bb}_{75}$ & AP$^\text{mk}$&   AP$^\text{mk}_{50}$ &  AP$^\text{mk}_{75}$&AP$^\text{bb}$&   AP$^\text{bb}_{50}$ &  AP$^\text{bb}_{75}$ & AP$^\text{mk}$&   AP$^\text{mk}_{50}$ &  AP$^\text{mk}_{75}$\\
	\midrule
	Supervised &100\%& 38.9& 59.6& 42.7& 35.4& 56.5& 38.1& 
   40.6& 61.3& 44.4 &36.8& 58.1& 39.5\\
    MoCo~\cite{moco} & 0\% & 38.5 & 58.9 & 42.0 & 35.1 &55.9 &37.7 &40.8&61.6&44.7&36.9&58.4&39.7 \\
	CoMatch&1\%&39.7& 61.2& 43.1& 36.1& 57.8& 38.5&41.2&62.2&44.9&37.3& 59.0&39.9\\	    
	CoMatch&10\%&\textbf{40.5}& \textbf{61.5}& \textbf{44.2}& \textbf{36.7}&  \textbf{58.3}&  \textbf{39.2}&
	\textbf{41.5}&\textbf{62.5}&\textbf{45.4}&\textbf{37.6}&\textbf{59.5}&\textbf{40.3}\\	       
	\bottomrule
	\end{tabular}
\vspace{-1ex}
	\caption
	{
		\small	
		Transfer the pre-trained models to object detection and instance segmentation on COCO, by fine-tuning Mask-RCNN with R50-FPN on \texttt{train2017}.
		We evaluate bounding-box AP (AP$^\text{bb}$) and mask AP (AP$^\text{mk}$) on \texttt{val2017}.
	}
	\label{tbl:coco}
	\vspace{-2ex}
\end{table*}			    				

%% file: sec_conclusion.tex
\vspace{-3ex}
\section{Conclusion}
\vspace{-0.5ex}
To conclude,
the success of CoMatch can be attributed to three contributions: (1) co-training of class probabilities and image embeddings, (2) memory-smoothed pseudo-labeling to mitigate confirmation bias, (3) graph-based contrastive learning to learn better representations.
We believe that CoMatch will help enable machine learning to be deployed in domains where labels are expensive to acquire.

%% file: sec_appendix.tex
\clearpage

\newpage		
\begin{appendices}

\section{Experiment Details}	
\label{sec:details}

In Table~\ref{tbl:hyperparameter}, we show the complete set of hyperparameters in our semi-supervised learning experiments.

\input{table/hyperparameter}

The strong augmentation $\mathrm{Aug_s}$ on ImageNet unlabeled data uses color distortion in addition to the standard crop-and-flip.
A pseudo-code for the color distortion in PyTorch is as follows:

\noindent\texttt{\small
from torchvision import transforms as T\newline
color\_jitter = T.ColorJitter(0.4,0.4,0.4,0.1)\newline
transforms.Compose([}

\texttt{\small 
	T.RandomApply([color\_jitter], p=0.8)
}

\texttt{\small 
	T.RandomGrayscale(p=0.2)])
}

\section{MB and MQ in CoMatch}
Figure~\ref{fig:mb} illustrates how the EMA model is utilized in CoMatch to construct the memory bank (MB) and the momentum queue (MQ).
The memory bank contains the class probability and the low-dimensional embeddings for both weakly-augmented labeled samples and weakly-augmented unlabeled samples.
The momentum queue contains the pseudo-labels for the unlabeled samples and their strongly-augmented embeddings.
 
\input{table/fig_mb_mq}

\section{Pseudo-code of CoMatch}	
Algorithm~\ref{alg} presents the pseudo-code of CoMatch.
\input{table/algorithm}

\end{appendices}

%% file: table/hyperparameter.tex
\begin{table}[!ht]
\small
\setlength\tabcolsep{2.5pt}
	\centering
	\resizebox{\columnwidth}{!}{
	\begin{tabular}	{l  | c  c c  c c c c  c c }
	\toprule
    Dataset   &$B$ & $\mu$ & $\lambda_{cls}$ &$\alpha$ & $K$ &  $t$&$\tau$  & $T$&$\lambda_{ctr}$\\
	\midrule
    \multirow{3}{*}{CIFAR-10} & \multirow{4}{*}{64} &  \multirow{4}{*}{7}  &  \multirow{4}{*}{1}  &  \multirow{4}{*}{0.9} &  \multirow{4}{*}{2560} & \multirow{4}{*}{0.2} & \multirow{4}{*}{0.95} &  \multirow{4}{*}{0.8}  & \multirow{3}{*}{1}\\
     & & & & & & & && \\
    & & & & & & & & &  \\
    STL-10   &&&&&  & & & &5\\
    \midrule
    ImageNet 1\% labels & \multirow{2}{*}{160}  & \multirow{2}{*}{4}  & \multirow{2}{*}{10} &\multirow{2}{*}{0.9} &\multirow{2}{*}{30000} &\multirow{2}{*}{0.1}  & 0.6 & 0.3 & 10\\
    ImageNet 10\% labels & & &&&&& 0.5& 0.2&2\\
	\bottomrule
	\end{tabular}	
 }
\vspace{-1ex}
	\caption
		{\small	
		 Hyperparameters for CoMatch in the semi-supervised learning experiments.
		}
	\vspace{-1ex}
	\label{tbl:hyperparameter}
\end{table}

%% file: table/fig_mb_mq.tex
\begin{figure}[h]
	\centering
	\includegraphics[width=\columnwidth]{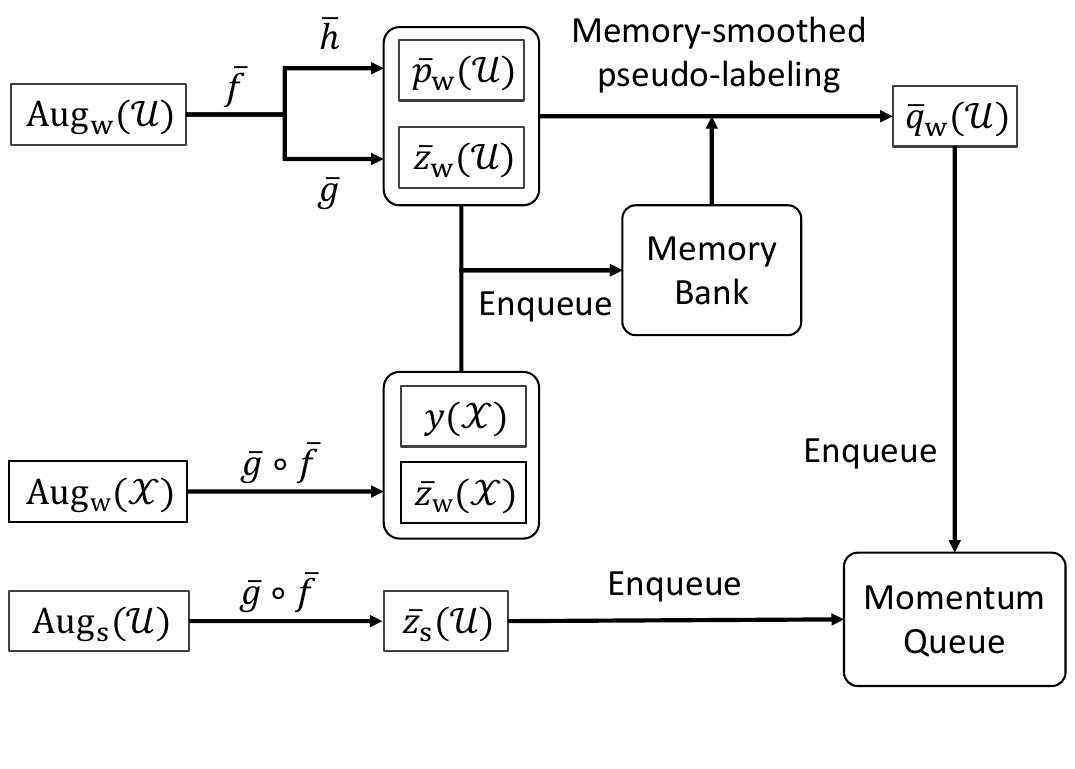}
  \vspace{-6.5ex}
  \caption
  	{ \small
	Illustration of the memory bank and the momentum queue.
	$\mathcal{U}$ is the batch of unlabeled data,
	$\mathcal{X}$ is the batch of labeled data.
	$\bar{f}$, $\bar{h}$, and $\bar{g}$ refer to the EMA version of the encoder, the classification head, and the projection head, respectively.
	  } 
  \label{fig:mb}
 \end{figure} 

%% file: table/algorithm.tex
\newcommand\mycommfont[1]{\footnotesize\ttfamily\textcolor{PineGreen}{#1}}
\SetCommentSty{mycommfont}

\begin{algorithm}[!h]
\small	
	\DontPrintSemicolon
	\SetNoFillComment
	
	\textbf{Input:} labeled batch $\mathcal{X}=\{(x_b,y_b)\}_{b=1}^B$, unlabeled batch $\mathcal{U}=\{u_b\}_{b=1}^{\mu B}$,  encoder $f$, classifier $h$, projection head $g$, memory bank $\mathrm{MB}=\{(p_k^w,z_k^w)\}_{k=1}^K$.\\

	\For {$b \in \{1,...,\mu B\}$}	
	{   \tcp{class probability prediction}
		$p_b^w=h\circ f(\mathrm{Aug_w}(u_b))$ \\
		\tcp{distribution alignment}
		$p_b^w=\text{DA}({p_b^w})$ \\
		 \tcp{weakly-augmented embedding}
		$z_b^w=g\circ f(\mathrm{Aug_w}(u_b))$ \\
		\tcp{memory-smoothed pseudo-labeling}
		\For {$k \in \{1,...,K\}$}	
		{
			$a_k = \frac{\exp({z^w_b}\cdot{z^w_k}/t)}{\sum_{k=1}^{K}\exp({z^w_b}\cdot{z^w_k}/t)}$\tcp*{affinity}
		}
		$q_b  = \alpha  p^w_b + (1-\alpha) \sum_{k=1}^K a_k p^w_k$ \\
		\tcp{strongly-augmented embeddings}
		$z_b=g\circ f(\mathrm{Aug_s}(u_b))$ \\
		$z'_b=g\circ f(\mathrm{Aug'_s}(u_b))$ \\		
	}
	\For {$b \in \{1,...,\mu B\}$}	
	{
		\For {$j \in \{1,...,\mu B\}$}		
		{
			\tcp{pseudo-label graph}	
			$W^q_{bj} = 
			\begin{cases}
			1 & \text{if}~b=j\\
			q_b \cdot q_j & \text{if}~b\neq j \text{~and~} q_b \cdot q_j\geq T\\
			0 & \text{otherwise}
			\end{cases}$\\
			\tcp{embedding graph}		
			$W^z_{bj} = 
			\begin{cases}
			\exp(z_b \cdot z'_b/t)& \text{if}~b=j\\
			\exp(z_b \cdot z_j/t)& \text{if}~b\neq j \\
			\end{cases}$
		}
	   $\hat{W}^q=\text{Normalize}(W^q)$\\
	   $\hat{W}^z=\text{Normalize}(W^z)$
}
\tcp{losses}
	$\mathcal{L}_x = \frac{1}{B} \sum_{b=1}^{B} \mathrm{H}(y_b,p(y|\mathrm{Aug_w}(x_b)))$\\
	$\mathcal{L}_u^{cls} = \frac{1}{\mu B} \sum_{b=1}^{\mu B} \mathbbm{1} (\max q_b\geq \tau) \mathrm{H}(q_b,p(y|\mathrm{Aug_s}(u_b)))$\\
	$\mathcal{L}_u^{ctr} = \frac{1}{\mu B} \sum_{b=1}^{\mu B} \mathrm{H}(\hat{W^q_b},\hat{W^z_b})$ \\
	$\mathcal{L}=\mathcal{L}_x + \lambda_{cls} \mathcal{L}_u^{cls} + \lambda_{ctr} \mathcal{L}_u^{ctr}$	\\
	update $f$, $h$, $g$ with SGD to minimize $\mathcal{L}$.
\caption{\small Pseudo-code of CoMatch (one iteration).}
\label{alg}
\end{algorithm}	 	